\documentclass{amsart}
\newcommand{\Real}{\mathbb R}
\renewcommand{\b}[1]{\mathbf{#1}}
\newcommand{\bx}{\b{x}}
\newcommand{\by}{\b{y}}
\newcommand{\bu}{\b{u}}

\begin{document}

\title{Kalman filter control in the reinforcement learning framework}%
\author{Istv\'{a}n Szita \and Andr\'{a}s L\H{o}rincz}
\address{Department of Information Systems \\ E\"{o}tv\"{o}s Lor\'{a}nd University of
Sciences \\ P\'azm\'any P\'eter s\'et\'any 1/C \\
1117 Budapest, Hungary}

\begin{abstract}
There is a growing interest in using Kalman-filter models in brain
modelling. In turn, it is of considerable importance to make
Kalman-filters amenable for reinforcement learning. In the usual
formulation of optimal control it is
computed off-line by solving a backward recursion. In this
technical note we show that slight modification of the
linear-quadratic-Gaussian Kalman-filter model allows the on-line
estimation of optimal control and makes the bridge to
reinforcement learning. Moreover, the learning rule for value
estimation assumes a Hebbian form weighted by the error of the
value estimation.
\end{abstract}


\maketitle

\section{Motivation}

Kalman filters and their various extensions are well studied and
widely applied tools in both state estimation and control.
Recently, there is an increasing interest in Kalman-filters or
Kalman-filter like structures as models for neurobiological
substrates. It has been suggested that Kalman-filtering (i) may
occur at sensory processing \cite{rao97dynamic,rao99predictive},
(ii) may be the underlying computation of the hippocampus, and may
be the underlying principle in control architectures
\cite{todorov02optimal,todorov02optimalsup}. Detailed
architectural similarities between Kalman-filter and the
entorhinal-hippocampal loop as well as between Kalman-filters and
the neocortical hierarchy have been described recently
\cite{lorincz00parahippocampal,lorincz02mystery}. Interplay
between the dynamics of Kalman-filter-like architectures and
learning of parameters of neuronal networks has promising aspects
for explaining known and puzzling phenomena, such as priming,
repetition suppression and categorization
\cite{lorincz02relating,keri02categories}.

As it is well known, Kalman-filter provides an on-line estimation
of the state of the system. On the other hand, optimal control
cannot be computed on-line, because it is typically given by a
backward recursion (the Ricatti-equations). For on-line parameter
estimations without control aspects, see \cite{rao99optimal}.

The aim of this paper is to derive an on-line control method for
the Kalman-filter and achieve optimal performance asymptotically.
Slight modification of the linear-quadratic-Gaussian (LQG)
Kalman-filter model is introduced for treating the LQG model as a
reinforcement learning (RL) problem.

\section{the Kalman filter and the LQG model}

Consider a linear dynamical system with state $\b x_t \in \Real^n$, control $\b u_t \in
\Real^m$, observation $\b y_t \in \Real^k$, noises $\b w_t \in \Real^n$ and $\b e_t \in
\Real^k$ (which are assumed to be Gaussian and white, with covariance matrix $\Omega^w$
and $\Omega^e$, respectively), in discrete time $t$:
\begin{eqnarray}
  \b x_{t+1} &=& F \b x_t + G \b u_t + \b w_t \\
  \b y_t &=& H \b x_t + \b e_t,
\end{eqnarray}
the initial state has mean $\hat{\b x}_1$ and covariance $\Sigma_1$. Executing the
control step $\b u_t$ in $\b x_t$ costs
\begin{equation}
c(\bx_t, \bu_t) := \bx_t^T Q \bx_t + \bu_t^T R \bu_t,
\end{equation}
and after the $N$th step the controller halts and receives a final cost of $\bx_N^T Q_N
\bx_N$.

This problem has the well known solution
\begin{eqnarray}
  \hat{\bx}_{t+1} &=& F \hat{\bx}_t + G \bu_t + K_t(\by_t - H\hat{\bx}_t) \label{x_estimate} \\
  K_t &=& F \Sigma_t H^T (H \Sigma_t H^T + \Omega^e)^{-1} \\
  \Sigma_{t+1} &=& \Omega^w + F \Sigma_t F^T - K_t H \Sigma_t A^T \label{Sigma_estimate}
\hskip2cm\textrm{(state estimation)}
\end{eqnarray}
and
\begin{eqnarray}
  \bu_t &=& -L_t \hat{\bx}_t  \\
  L_t &=& (G^T S_{t+1} G + R)^{-1} G^T S_{t+1} F \\
  S_t &=& Q_t + F^T S_{t+1} F - F^T S_{t+1} G L_t.
\hskip2cm\textrm{(optimal control)}
\end{eqnarray}

Unfortunately, the optimal control equations are not on-line, because they can be solved
only by stepping backward from the final, $N$th step.

\section{Kalman Filtering in the Reinforcement Learning Framework}

First of all, we slightly modify the problem: the run time of the controller will not be
a fixed number $N$. Instead, after each time step, the process will be stopped with some
fixed probability $p$ (and then the controller incurs the final cost $c_f(\bx_f) :=
\bx_f^t Q_f \bx_f$).

\subsection{The cost-to-go function}

Let $V^*_t(\bx)$ be the optimal cost-to-go function at time step $t$, i.e.
\begin{equation}
  V^*_t(\bx) := \inf_{\bu_t, \bu_{t+1}, \ldots} E\bigl[ c(\bx_t,\bu_t) + c(\bx_{t+1},\bu_{t+1}) +
  \ldots + c_f(\bx_f)  \big| \bx_t = \bx \bigr].
\end{equation}
Clearly, for any $\bx$,
\begin{equation}
  V^*_t(\bx) = p \cdot c_f(\bx) + (1-p)\cdot  \inf_{\bu} \Bigl( c(\bx,\bu) +
   E_w \bigl[V^*_{t+1} (F\bx + G\bu +w)\bigr] \Bigr)
\end{equation}
It can be easily shown that the optimal cost-to-go function is time-independent,
furthermore, it is a quadratic function of $\bx$, that is, it is of the form
\begin{equation}
  V^*(\bx) = \bx^T \Pi^* \bx.
\end{equation}
Our task is to estimate $V^*$ (in fact, the parameter matrix $\Pi^*$) on-line. This will
be done by value iteration.

\subsection{Value iteration, greedy action selection and the temporal differencing error}

Value iteration starts with an arbitrary initial cost-to-go function $V_0(\bx) = \bx^T
\Pi_0 \bx$. After this, control actions are selected according to the current value
function estimate, the value function is updated according to the experience, and these
two steps are iterated.

The $t$th estimate of $V^*$ is $V_t(\bx) = \bx^T \Pi_t \bx$. The greedy control action
according to this is given by
\begin{eqnarray}
  \bu_t &=& \arg\min_{\bu} \Bigl( c(\bx_t,\bu)+ E \bigl[V_t(F \bx_t + G \bu + w)\bigr] \Bigr) \\
      &=& \arg\min_{\bu} \Bigl( \bu^T R \bu + (F \bx_t + G \bu)^T \Pi_t (F \bx_t + G \bu) \Bigr) \\
      &=& -(R+G^T \Pi_t G)^{-1} (G^T \Pi_t F) \bx_t.  \label{u_t}
\end{eqnarray}

For the sake of simplicity, the cost-to-go function will be
updated by using the 1-step temporal differencing (TD) method.
Naturally, it can be substituted with more sophisticated methods
like multi-step TD or eligibility traces. The TD error is
\begin{equation}
  \delta_t =
  \begin{cases}
    V_t(\bx_t) - c_f(\bx_t)  & \text{if the controller was stopped at the $t$th time step}, \\
    \bigl( c(\bx_t,\bu_t) + V_t(\bx_{t+1}) \bigr) - V_t(\bx_t), & \text{otherwise}.
  \end{cases}
\end{equation}
and the update rule for the parameter matrix $\Pi_t$ is
\begin{eqnarray}
  \Pi_{t+1} &=& \Pi_t + \alpha_t \cdot \delta_t \cdot \nabla_{\Pi_t} V_t(\bx_t) \\
            &=& \Pi_t + \alpha_t \cdot \delta_t \cdot \bx_t \bx_t^T,  \label{Pi_update}
\end{eqnarray}
where $\alpha_t$ is the learning rate. Note that value-estimation
error weighted Hebbian learning rule has emerged.

\section{Concluding remarks}

The Kalman-filter control problem was slightly modified to fit
the RL framework and an on-line control rule was
achieved. The well-founded theory of reinforcement learning
ensures asymptotic optimality for the algorithm. The described
method is highly extensible. There are straightforward
generalizations to other cases, e.g., to extended Kalman filters,
dynamics with unknown parameters, non-quadratic cost functions, or
more advanced RL algorithms, e.g. eligibility traces. For
quadratic loss functions, we have found that learning is Hebbian and it
is weighted by the error of value-estimation.

\bibliographystyle{amsplain}
\bibliography{RL_KF}

\providecommand{\bysame}{\leavevmode\hbox to3em{\hrulefill}\thinspace}
\providecommand{\MR}{\relax\ifhmode\unskip\space\fi MR }
% \MRhref is called by the amsart/book/proc definition of \MR.
\providecommand{\MRhref}[2]{%
  \href{http://www.ams.org/mathscinet-getitem?mr=#1}{#2}
}
\providecommand{\href}[2]{#2}
\begin{thebibliography}{1}

\bibitem{keri02categories}
Sz. K\'eri, Gy. Benedek, Z.~Janka, P.~Aszal\'os, B.~Szatm\'ary, G.~Szirtes, and
  A.~L{\H o}rincz, \emph{Categories, prototypes and memory systems in
  alzheimer's disease}, Trends in Cognitive Science \textbf{6} (2002),
  no.~132-136.

\bibitem{lorincz00parahippocampal}
A.~L{\H o}rincz and G.~Buzs\'aki, \emph{The parahippocampal region:
  \mbox{I}mplications for neurological and psychiatric dieseases}, Annals of
  the New York Academy of Sciences (H.E. Scharfman, M.P. Witter, and
  R.~Schwarz, eds.), vol. 911, New York Academy of Sciences, New York, 2000,
  pp.~83--111.

\bibitem{lorincz02mystery}
A.~L{\H o}rincz, B.~Szatm\'ary, and G.~Szirtes, \emph{Mystery of structure and
  function of sensory processing areas of the neocortex: \mbox{A} resolution},
  J. Comp. Neurosci. \textbf{13} (2002), 187–205.

\bibitem{lorincz02relating}
A.~L{\H o}rincz, G.~Szirtes, B.~Tak\'acs, I.~Biederman, and R.~Vogels,
  \emph{Relating priming and repetition suppression}, Int. J. of Neural Systems
  \textbf{12} (2002), 187--202.

\bibitem{rao99optimal}
R.P.N. Rao, \emph{An optimal estimation approach to visual perception and
  learning}, Vision Research \textbf{39} (1999), 1963--1989.

\bibitem{rao97dynamic}
R.P.N. Rao and D.H. Ballard, \emph{Dynamic model of visual recognition predicts
  neural response properties in the visual cortex}, Neural Comput \textbf{9}
  (1997), 721--763.

\bibitem{rao99predictive}
\bysame, \emph{Predictive coding in the visual cortex: \mbox{A} functional
  interpretation of some extra-classical receptive-field effects}, Nature
  Neuroscience \textbf{2} (1999), 79--87.

\bibitem{todorov02optimal}
E.~Todorov and M.I. Jordan, \emph{Optimal feedback control as a theory of motor
  coordination}, Nature Neuroscience \textbf{5} (2002), 1226--1235.

\bibitem{todorov02optimalsup}
\bysame, \emph{Supplementary notes for optimal feedback control as a theory of
  motor coordination}, Nature Neuroscience website, 2002.

\end{thebibliography}

\end{document}